\documentclass{article}

\usepackage{arxiv}

\usepackage[utf8]{inputenc} 
\usepackage[T1]{fontenc}    
\usepackage{textcomp}       

\usepackage{hyperref}       
\usepackage{url}            
\usepackage{booktabs}       
\usepackage{amsfonts}       

\usepackage{graphicx}

\usepackage{amsmath}

\usepackage{subfig}
\usepackage{color}
\usepackage{float}
\usepackage{tabularx}
\usepackage{appendix}
\newcolumntype{C}{>{\centering\arraybackslash}X}

\title{Oldies but Goldies: The Potential of Character N-grams for Romanian Texts}

\author{
 Dana Lup\c sa \\
  Babeș-Bolyai University \\
  \texttt{dana.lupsa@ubbcluj.ro} \\
 \And
 Sanda-Maria Avram \\
  Babeș-Bolyai University \\
  \texttt{sanda.avram@ubbcluj.ro} \\
   \And
  Radu Lup\c sa \\
  Babeș-Bolyai University \\
  \texttt{radu.lupsa@ubbcluj.ro} \\
}

\begin{document}
\maketitle



 \begin{abstract}
This study addresses the problem of authorship attribution for Romanian texts using the ROST corpus, a standard benchmark in the field. We systematically evaluate six machine learning techniques—Support Vector Machine (SVM), Logistic Regression (LR), k-Nearest Neighbors (k-NN), Decision Trees (DT), Random Forests (RF), and Artificial Neural Networks (ANN), employing character n-gram features for classification. Among these, the ANN model achieved the highest performance, including perfect classification in four out of fifteen runs when using 5-gram features. These results demonstrate that lightweight, interpretable character n-gram approaches can deliver state-of-the-art accuracy for Romanian authorship attribution, rivaling more complex methods. Our findings highlight the potential of simple stylometric features in resource-constrained or under-studied language settings.
\end{abstract} 

\bigskip

\section{Introduction}
The authorship determination of a text requires distinguishing the author particularities that are reflected in the written text. Instruments used to do automated authorship attribution (AA) can be characterized along several dimensions: 

\begin{itemize}
    \item \textit{Dataset characteristics:} including language, dataset size, class imbalance, and text type (e.g., emails, novels, social media posts, or cross-domain/genre scenarios)~\cite{stamatatos2009survey};
    \item \textit{Preprocessing steps:} 
    include text normalization and punctuation handling~\cite{potthast2011cross};
    \item \textit{Feature extraction:} encompassing character-based features (e.g., character types - letters, digits; character n-grams), lexical features (e.g., word frequencies, word n-grams, stop words, function words), syntactic features (e.g., parts of speech, sentence and phrase structure), and semantic features (e.g., topic modeling, word embeddings, contextual representations from large language models)~\cite{kestemont2014function,neal2017surveying,stamatatos2009survey,ngram2022};
   
    \item \textit{Computational methods:} ranging from traditional statistical techniques (e.g., stylometry, 
    PCA)
    to machine learning algorithms (e.g., Support Vector Machines, Random Forests), deep learning models (e.g., Recurrent Neural Networks, transformers), and modern large language model (LLM) approaches~\cite{koppel2009computational}; 
    \item \textit{Evaluation metrics:} including accuracy, macro-averaged accuracy (especially for imbalanced datasets), precision, recall, F1-score, area under the curve, and confusion matrices~\cite{stamatatos2009survey}.
\end{itemize}

English remains the most studied language in AA research due to the availability of data and the focus on computational linguistics~\cite{stamatatos2009survey, kestemont2014function}. However, research on Romanian texts has seen growing interest in recent years, both by efforts to support under-resourced languages and by practical applications such as historical and literary analysis, plagiarism detection, and cybercrime investigations~\cite{dinu2008authorship,avram2022comparison,avram2023bert,nitu2024authorship}.

The size of the dataset significantly influences the choice of methodology. Deep learning and LLM-based approaches typically require large datasets for effective training, validation, and testing. In contrast, classical machine learning methods are often more suitable for small to mid-scale datasets (e.g., 10,000–20,000 texts, tens of authors)~\cite{kestemont2018overview}.

Key challenges associated with AA datasets include~\cite{stamatatos2013robustness, neal2017surveying}: 
\begin{itemize}
    \item limited text availability (e.g., few samples per author),
    \item class imbalance (e.g, 
    in representation of authors or genres),
    \item generalizability issues (e.g., 
    on cross-domain or cross-genre texts).
\end{itemize}

The aforementioned challenges are are especially pronounced for under-resourced languages like Romanian~\cite{avram2022comparison,avram2023bert, nitu2024authorship}.

Dataset partitioning into training, testing, and validation subsets also affects performance outcomes, as different splits can yield varying results due to random sampling effects. 
Thus, researchers commonly employ multiple splits or cross-validation techniques to obtain more reliable performance estimates.

While preprocessing transforms raw text into a cleaner format, the feature representation stage, where texts are converted to numerical vectors, is critical to classification success. The choice of features can have a profound impact on the results, often independent of the classification algorithm used, making feature engineering a central focus of AA research.

Different classification algorithms may yield varying results on the same data and features. Consequently, studies often compare multiple algorithms or maintain algorithm consistency when evaluating features. Furthermore, optimizing algorithm hyperparameters can significantly enhance performance.

In this paper, we investigate the effectiveness of character n-gram features to improve authorship attribution performance on the Romanian ROST dataset.

The remainder of this paper is organized as follows. Section~\ref{related-work} reviews related work on authorship attribution, with a focus on approaches relevant to Romanian texts and the ROST dataset. Section~\ref{sec:experiments} presents the experimental setup and results, including dataset description, model configurations, and detailed analyses of classification performance across different feature and parameter settings. Finally, Section~\ref{sec:conclusions} concludes the paper with a summary of key findings and outlines directions for future research.

\section{Related Work}\label{related-work}
\subsection{Challenges in Romanian Authorship Attribution}

Authorship attribution (AA) in under-resourced languages. such as Romanian, presents a distinct set of challenges that set it apart from work in high-resource languages. One of the primary obstacles is the limited availability of annotated corpora, which restricts both the scale and diversity of training data available for model development and evaluation. This scarcity is compounded by a significant class imbalance, as datasets such as the ROST~\cite{avram2022comparison} corpus exhibit uneven representation of authors, ranging from as few as 27 to as many as 60 texts per author, and substantial variation in text length, from short stories of 90 words to extensive works exceeding 39,000 words. The ROST dataset also encompasses a wide range of genres, including stories, fairy tales, novels, articles, and sketches, and spans a broad historical period from 1850 to 2023, further increasing heterogeneity and complicating model generalization.

These factors introduce biases and variability that are difficult to control, making it challenging to develop robust and generalizable AA models for Romanian~\cite{avram2022comparison, avram2023bert}. Addressing such issues is essential for advancing the field, as models trained on unbalanced or limited data may fail to perform reliably across different authors, genres, or time periods.

Recent research has begun to address these challenges by developing hybrid models that combine handcrafted linguistic features with contextualized embeddings, tailored specifically for Romanian and other under-resourced languages. However, the linguistic complexity of Romanian, characterized by rich morphology and flexible word order, means that effective solutions for English or other major languages often do not transfer directly~\cite{nitu2024authorship}. Similar difficulties have been reported in other under-resourced languages, such as Albanian, where the lack of large annotated corpora continues to impede progress in authorship attribution research~\cite{misini2024automatic}.

Moreover, the scarcity of large, publicly available Romanian corpora limits the applicability of data-intensive deep learning methods, making careful feature engineering and rigorous evaluation protocols especially important for achieving reliable results in this context.

\subsection{N-gram Features in Authorship Attribution} 
\label{sec:NgramFeat}

N-grams are contiguous sequences of $N$ items, such as characters, words, or other tokens, extracted from text to capture 
patterns indicating an author's unique style. Typically, n-grams are constructed at the character, word, or syntactic level \cite{ngram2022}.

Character n-grams have been extensively used in authorship attribution research due to their ability to encode stylistic nuances, lexical patterns, word order tendencies, and punctuation or capitalization habits \cite{stamatatos2009survey}. Despite their widespread use in languages such as English, Romanian texts remain relatively understudied using this approach \cite{stamatatos2009survey, avram2022comparison}.

\subsubsection{Advantages of Character N-grams}
\begin{itemize}
    \item \textbf{Robustness:} 
    Character n-grams are resilient to infrequent errors such as grammatical mistakes or punctuation slips because their discriminative power derives 
    from frequent, recurring patterns \cite{stamatatos2013robustness}.
    \item \textbf{Preservation of Stylistic Traits:} 
    Subtle author-specific variations, such as repeated punctuation marks or distinctive capitalization, are naturally encoded within character n-grams~\cite{koppel2009computational}. For instance, some authors rarely use exclamation marks, while others employ them frequently. Similarly, preferences for sentence length can be reflected through punctuation usage patterns, such as the relative frequency of periods versus commas, which correspond to shorter or longer sentence structures, respectively~\cite{Howedi2014}.
    \item \textbf{Language Independence:} 
    Character n-grams do not require deep linguistic knowledge of grammar or semantics, making them applicable across diverse languages without modification.
    \item \textbf{Computational Efficiency:} 
    Extracting character n-grams is both 
    straightforward and computationally inexpensive, as it involves direct processing of raw text without the need for complex preprocessing.
\end{itemize}

\subsubsection{Limitations of Character N-grams}
\begin{itemize}
    \item \textbf{Redundancy:} 
    Due to the overlapping nature of n-grams, each n-gram shares $N-1$
    characters with adjacent ones. Therefore, many n-grams represent slight variations of the same lexical unit 
    (e.g., "\texttt{in\_}", "\texttt{in.}", "\texttt{in!}"). 
    While this redundancy can reinforce stylistic signals such as affixation (prefix/suffix) or punctuation preferences, it may also lead to overfitting if not properly managed.
    \item \textbf{Sparsity and High Dimensionality:}   
    Compared to word-level n-grams, character n-grams, especially for larger $N$, tend to generate very high-dimensional and sparse feature spaces, which can pose challenges for model training and generalization \cite{stamatatos2013robustness, sapkota2015not}.
\end{itemize}

\subsection{Computational Methods}
\label{subsec:Algorithms}
A wide range of 
classification algorithms has been applied to authorship attribution (AA) leveraging character n-gram features~\cite{boser1992training,fix1951discriminatory,fix1952discriminatory,altman1992introduction,quinlan1986induction,zurada1992introduction}. Next, we provide a concise overview of the principal methods utilized in this study:

\textbf{Support Vector Machine (SVM)} 
is a supervised learning model that identifies an optimal hyperplane to separate author classes by maximizing the margin between data points of different classes. It excels in handling high-dimensional feature spaces such as those generated by n-grams and is robust against overfitting and noise, making it a popular choice in stylometric analysis.

\textbf{Logistic Regression (LR)} %
is a probabilistic linear model used for binary and multiclass classification. It maps input features to class probabilities via the logistic (sigmoid) function and applies regularization (L1 or L2 penalties) to prevent overfitting. LR provides interpretable coefficients, which can be valuable for understanding the contribution of specific n-grams, particularly in smaller datasets.

\textbf{k-Nearest Neighbors (k-NN)} 
is a non-parametric, instance-based classifier that assigns a class label based on the majority vote of the $k$ nearest neighbors in the feature space. It relies on distance metrics such as Euclidean, cosine, or Minkowski distance. While simple to implement and intuitive, k-NN can be computationally expensive for large datasets and high-dimensional n-gram features.

\textbf{Decision Trees (DT)} 
recursively partition the feature space by selecting feature thresholds that maximize class purity, typically using criteria like entropy or Gini impurity. The resulting tree structure yields interpretable decision rules that can highlight stylometric patterns, such as frequent character sequences. However, DTs are prone to overfitting, especially with high-dimensional n-gram data.

\textbf{Random Forest (RF)} 
is an ensemble method that constructs multiple decision trees using random subsets of the data and features, aggregating their predictions via majority voting. This approach reduces overfitting through bagging and feature randomness. RF handles noisy, high-dimensional n-gram data effectively and provides measures of feature importance, enabling identification of discriminative n-grams.

\textbf{Artificial Neural Networks (ANN)} 
consist of interconnected layers of neurons that learn hierarchical feature representations through backpropagation. Activation functions such as ReLU enable the modeling of complex, non-linear relationships among features. While deep ANNs typically require large datasets, shallow architectures (1–2 hidden layers) are well-suited for moderate-sized corpora like ROST~\cite{avram2022comparison}.

Among the aforementioned methods, SVM and RF have traditionally dominated AA research \cite{stamatatos2013robustness} due to their strong performance with stylometric features such as n-grams. ANNs, particularly deep learning models, are gaining traction but generally demand larger datasets. Logistic Regression and k-NN often serve as baseline benchmarks, while standalone Decision Trees are less commonly employed due to their susceptibility to overfitting.

\subsection{Evaluation Metrics}
\label{subsec:metrics}
Authorship attribution studies commonly employ the following evaluation metrics to assess model performance:

\subsubsection{Accuracy} 
Accuracy measures the proportion of correctly classified texts across all classes. 
 While widely used, accuracy can be misleading for imbalanced datasets, as it may overemphasize performance on majority classes.

\subsubsection{Macro-Accuracy (Balanced Accuracy)} 
Macro-accuracy addresses class imbalance by calculating accuracy for each class independently and then averaging the results:  
$$
\text{Macro-Accuracy} = \frac{1}{C} \sum_{i=1}^{C} \text{Accuracy}_{Class_i}
$$  
where $C$ is the number of classes (authors). This metric ensures all authors contribute equally to the final score, making it more representative for datasets with uneven class distributions.

\subsubsection{Precision, Recall, and F1-Score}
These metrics provide complementary insights into model performance, particularly for imbalanced data:  
\begin{itemize}
    \item \textbf{Precision} 
    The proportion of correctly attributed texts 
    for an author out of all texts predicted as that author. 
  High precision minimizes false attributions. 
  
    \item \textbf{Recall} 
    The proportion of correctly attributed texts for an author out of all texts actually written by that author. 
  High recall indicates strong retrieval of true positives.  

    \item \textbf{F1-Score} The harmonic mean of precision and recall. It reflects how well a system is at finding relevant items (precision) and all of them (recall), without being skewed by abundant irrelevant items. 
    Particularly important for imbalanced datasets, where, if one class significantly outnumbers the others, a model can achieve high accuracy by simply predicting the majority class.
    
\end{itemize}

\subsubsection{Confusion Matrix} A table showing the number of correct and incorrect predictions for each class. Reveals which authors are frequently confused. It summarizes the true positives, true negatives, false positives, and false negatives, from which accuracy is derived.

\section{Experiments and results}
\label{sec:experiments}

\subsection{The dataset}
Motivated by the relative scarcity of research on Romanian authorship attribution (AA) and the challenges posed by the ROST corpus, our work aims to advance the field by exploring lightweight yet effective feature representations. The ROST dataset, comprising approximately 400 texts authored by 10 writers, is characterized by significant class imbalance and diverse text lengths and genres, making it a challenging benchmark for AA~\cite{avram2022comparison, avram2023bert}.

Previous studies have established important baselines on ROST:  
\begin{itemize}
    \item \textbf{Avram et al. (2022)} 
    introduced the dataset and evaluated five classification methods, including Artificial Neural Networks (ANN), Multi Expression Programming (MEP), k-Nearest Neighbors (k-NN), Support Vector Machines (SVM), and C5.0 Decision Trees, using inflexible part-of-speech tags as features. Their best macro-accuracy was 80.94\%, with an overall error rate of 20.40\%~\cite{avram2022comparison}.
    
    \item \textbf{Avram (2023)} 
    leveraged a Romanian-specific BERT model, 
    achieving approximately 87\% macro-accuracy~\cite{avram2023bert}.  
    
    \item \textbf{Nitu et al. (2024)} 
    proposed a hybrid Transformer architecture combining handcrafted linguistic features (lexical, syntactic, semantic, discourse markers) with BERT embeddings, reaching a state-of-the-art F1-score of 0.95 and reducing the error rate to 4\%~\cite{nitu2024authorship}.
\end{itemize}

While transformer-based approaches demonstrate impressive performance, their reliance on large pretrained models and substantial computational resources limits their accessibility, especially in resource-constrained environments. In contrast, our study investigates character n-grams, a lightweight, interpretable feature set, to assess whether they can achieve comparable performance on the ROST dataset. By systematically evaluating n-gram sizes ranging from 2 to 5, we aim to demonstrate that simpler, computationally efficient methods can provide competitive accuracy while offering greater interpretability and ease of deployment.

\subsection{Data Preprocessing}
In the raw text preprocessing phase, we apply \emph{normalization} to standardize the textual data. Therefore, texts were initially preprocessed to address specific character encoding variations like:
\begin{itemize}
    \item \textbf{Diacritic standardization:} converting 
    Romanian characters such as Ş/ş and Ţ/ţ to their canonical forms Ș/ș and Ț/ț;
    \item \textbf{Punctuation unification:}
        \begin{itemize}
            \item convert smart quotes 
            (
            „ “ ” ’
            ) 
            to straight quotes (\texttt{"})
            \item convert en/em dashes (–—―) to hyphens (\texttt{-})
            \item convert ellipses (…) to triple periods (\texttt{...})
        \end{itemize}
    \item \textbf{Whitespace regularization:} collapse multiple contiguous whitespace into a single spaces
\end{itemize}
These steps ensure consistency and reduce noise in the dataset, facilitating more reliable downstream analysis.

In addition to normalization, we performed the following preprocessing steps:
\begin{itemize}
    \item \textbf{Case Handling:} We conducted experiments using both lowercase text and the original case, to assess the impact of letter casing on authorship attribution.
    \item \textbf{Digit Replacement:} Since the specific values of the numbers were not relevant to our analysis, all digits were replaced with a special character (@), which does not occur in the original texts.
    \item \textbf{Punctuation Preservation:} All other special characters (punctuation marks) were left unchanged, as patterns in punctuation usage may reflect individual authorial style. 
    \item \textbf{Whitespace Encoding:} Recognizing the potential importance of whitespace as a stylistic marker, we replaced spaces and tabs with the underscore character (\_), and newlines with the dollar sign (\$). This allows us to explicitly encode paragraph boundaries and whitespace usage into the N-grams as distinct features.
\end{itemize}

\subsection{Feature Extraction Process}
We employ character N-grams to construct stylometric representations of texts, following these steps:

\subsubsection{N-gram Definition and Range.} Drawing on prior stylometric studies (Houvardas et al., 2006; Ramezani et al., 2013; Ntoulas et al., 2006; Smith \& Jones, 2022)~\cite{Houvardas2006,2013Ramezani,ngram-contin2006,ngram2022}, we analyze character sequences of length $N=2$ to $N=5$. This range captures both local orthographic patterns (e.g., bigrams) and longer morphological features (e.g., pentagrams).

\subsubsection{Vectorization via TF-IDF} Each document is transformed into a numerical vector using Term Frequency-Inverse Document Frequency (TF-IDF) weighting. This approach emphasizes N-grams that are statistically distinctive to individual documents while down-weighting common sequences.

\subsubsection{Comprehensive Feature Inclusion.} To avoid potential information loss from premature feature selection, we retain all extracted character N-grams during initial modeling. This ensures maximal preservation of potential stylistic markers.

\subsubsection{Feature Matrix Construction.} The final representation is a 
$D \times F$ matrix, where:
\begin{itemize}
    \item $D =$ Number of documents
    \item $F =$ Total unique N-grams across all $N$ values
    \item Cell values = TF-IDF weights for 
    N-gram/document pairs
\end{itemize}

\subsection{Experimental Setup and Classification Methods}

\subsubsection{Implementation Details}
All experiments were conducted in Python, utilizing the following libraries and tools:
\begin{itemize}
    \item \textbf{Data Processing and Manipulation:} \texttt{numpy} and \texttt{pandas} for numerical operations and data handling.
    \item \textbf{Feature Extraction:} 
    using \texttt{TfidfVectorizer}  from \texttt{scikit-learn} to generate
    character N-gram feature representations.
    \item \textbf{Model Training and Evaluation:} \texttt{scikit-learn} for implementing and evaluating machine learning classifiers. 
\end{itemize}

\subsubsection{Classification Algorithms}
We explored several supervised learning algorithms:
Support Vector Machine (SVM), Logistic Regression (LR), k-Nearest Neighbor (k-NN), Decision Trees (DT), Random Forest (RF), and Artificial Neural Networks (ANN).

\subsubsection{Model Evaluation}
We use standard metrics including \emph{accuracy}, \emph{macro-accuracy} (aka. balanced accuracy or macro-averaged accuracy), and the \emph{classification report} (precision, recall, F1-score), as provided by \texttt{scikit-learn}.

\subsection{Experimental Procedure}
For each of the classification methods described above,
we conducted a set of experiments to identify the configuration yielding the best performance. Our investigation included the following key aspects: 
\begin{itemize}
    \item \textbf{Parameter Exploration:} For 
    k-NN, 
    we evaluated five different values of k to determine the optimal neighborhood size.
    \item \textbf{Robustness to Randomness:} To mitigate the effects of stochastic variability inherent in certain models, we repeated the training and evaluation of 
    DT, 
    RF, and 
    ANN 
    five times, each with a different random seed. This procedure allowed us to assess the stability and reliability of the results.
    \item \textbf{Evaluation Metrics:} Performance was assessed using multiple metrics, including accuracy and balanced accuracy, to provide a comprehensive understanding of classifier effectiveness across potentially imbalanced classes.
    \item \textbf{Data Splitting:} We employed randomly selected train-test splits to preserve class distribution during evaluation, ensuring fair and representative performance estimates.
\end{itemize}

This rigorous experimental protocol ensures that observed performance differences reflect genuine model capabilities rather than artifacts of data sampling or initialization.

\subsection{Parameter Tuning and Model Selection}

\begin{table}[tb]
\centering
\footnotesize
\begin{tabular}{lll}
\textbf{ML Model} & \textbf{Hyperparameters} & \textbf{Variation Parameters} \\
\toprule
SVM & kernel: linear & — \\
LR & solver: lbfgs, penalty: l2 & — \\
k-NN & metric: minkowski & n\_neighbors  $=\{3, 5, 7, 9, 11\}$ \\
DT & criterion: gini & randomstate  $=\{7, 17, 42, 67, 101\}$ \\
RF & n\_estimators: 100 & randomstate $=\{7, 17, 42, 67, 101\}$ \\
 &  criterion: gini & \\

ANN & activation: relu & randomstate  $=\{7, 17, 42, 67, 101\}$ \\
              & hidden layer sizes: (100,50) &  \\
\bottomrule
\end{tabular}
\caption{\footnotesize Summary of machine learning models and their hyperparameters. To ensure that performance differences reflect model characteristics rather than random variation, 
DT, RF, ANN were each trained and evaluated five times using different random seeds. For k-Nearest Neighbors (k-NN), five different values of the number of neighbors were tested.}
\label{tab:ML_parameters}
\end{table}

The dataset was partitioned into training (80\%) and testing (20\%) subsets. This splitting procedure was repeated five times to generate distinct train-test partitions, ensuring robustness and reliability of the evaluation. For each partition, all six machine learning models (SVM, LR, k-NN, DT, RF, ANN) were trained and evaluated using the hyperparameter configurations detailed in Table~\ref{tab:ML_parameters}.

Furthermore, to assess the impact of text casing on model performance, experiments were conducted on both the original case and fully lowercased versions of the texts.

\subsection{Results Overview and Key Findings}

\subsubsection{Impact of Letter Casing on Classification Performance}

\begin{figure}[tb] 
\centering
\includegraphics[width=0.7\textwidth]{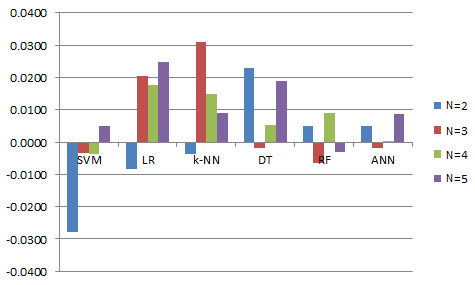}
\caption{\footnotesize 
Differences in Macro-Accuracy (MAcc) between fully lowercased and original-case texts across n-gram sizes ($N=2$ to $5$) for six classification methods: SVM, LR, k-NN, DT, RF, and ANN. The values used for the differences were the average of the macro-accuracy scores computed over multiple experimental runs.}
\label{fig:diff_lower-orig}
\end{figure}

We conducted experiments comparing classification performance on original-case versus fully lowercased texts. The observed differences in Macro-Accuracy (MAcc) were generally minor. The largest difference, 0.031, occurred with the k-Nearest Neighbors (k-NN) classifier at n-gram size $N=3$, where the lowercase representation slightly outperformed the original case.
 
Figure~\ref{fig:diff_lower-orig} depicts the variation in MAcc differences across all classifiers and n-gram sizes. Overall, lowercase texts tend to yield marginally better results, particularly for $k > 3$ and for the k-NN and 
LR 
classifiers. However, in most cases, the difference remains below 0.02, which is negligible and likely falls within the range of normal variation due to hyperparameter tuning.

For ANN, 
the average difference between casing strategies is even smaller, approximately 0.009. Interestingly, at n-gram size $N=5$, the trend reverses slightly, with original-case texts performing better, as further illustrated in Figure~\ref{fig:ANN_all15}.

These findings suggest that letter casing has no consistent or substantial impact on classification performance across the evaluated methods. The minor differences observed are unlikely to affect practical outcomes or model selection decisions.
 
 \subsubsection{The Impact of N-gram Size on Classification Performance}

\begin{figure} [tb] 
    \centering
    \subfloat[\centering]{\includegraphics[width=0.48\textwidth]{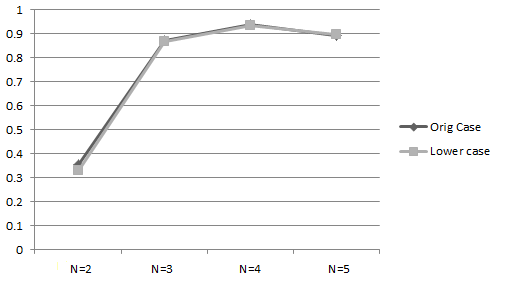}}
    \subfloat[\centering]{\includegraphics[width=0.48\textwidth]{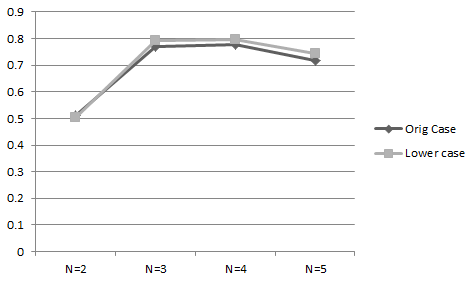}}
    
    \subfloat[\centering]{\includegraphics[width=0.48\textwidth]{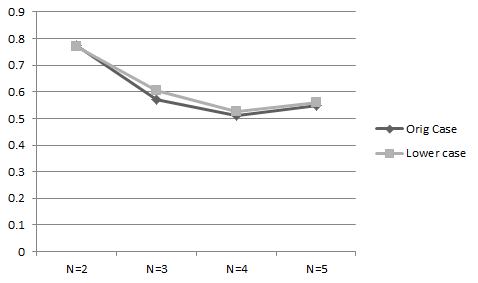}}
    \subfloat[\centering]{\includegraphics[width=0.48\textwidth]{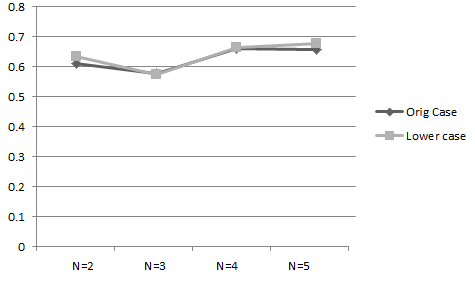}}
    
    \subfloat[\centering]{\includegraphics[width=0.48\textwidth]{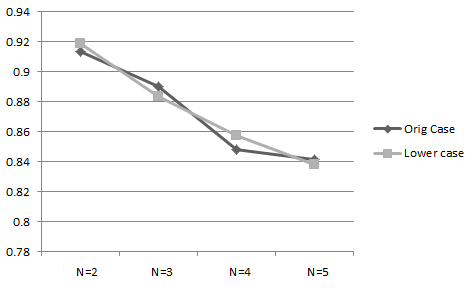}
        \label{fig:RFallKgr}}
    \subfloat[\centering]{\includegraphics[width=0.45\textwidth]{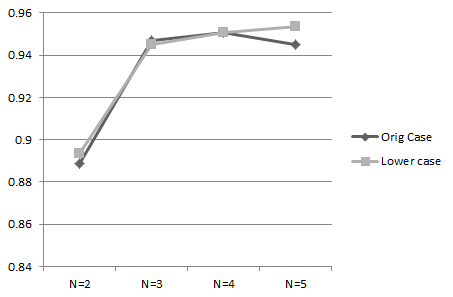}        \label{fig:ANNallKgr} 
    }
   
   \caption{\footnotesize
   Variation of MAcc 
   across different N-gram sizes ($N=2$ to $5$) for all six classification methods: (A) SVM, (B) LR, (C) k-NN, (D)  DT, (E) RF, and (F) ANN. 
   This analysis examines how varying the N-gram size affects the models’ performance. Values shown represent the average of the macro-accuracy scores computed over multiple experimental runs.  
   }
    \label{fig:allKgr5models}
\end{figure}

Figure~\ref{fig:allKgr5models} presents the variation in of Macro-Accuracy (MAcc) with respect to the N-gram size parameter $k$ (ranging from 2 to 5) for each of the six classification methods: SVM, LR, k-NN, DT, RF, and ANN.

It is important to note that in Figures~\ref{fig:RFallKgr} and \ref{fig:ANNallKgr}, the y-axis scale is adjusted to focus on the observed variations in MAcc, rather than spanning the full range from 0 to 1. This scaling facilitates a clearer visualization of performance trends as $k$ changes.

A noticeable decline in MAcc is observed for RF 
(Figure~\ref{fig:RFallKgr}) as the N-gram size increases. Conversely,  ANN 
shows a general upward trend in MAcc with larger N-gram dimensions. For SVM, 
the highest MAcc is achieved at $k=4$, while k-NN 
attains its peak performance at $k=2$. The remaining methods do not exhibit a clear monotonic trend in MAcc relative to the N-gram size.
 
\subsubsection{Best Results}
\label{sec:bestresults}

ANN 
achieved the best overall performance, with average accuracy and macro-accuracy (MAcc) values exceeding 0.9, as summarized in Table~\ref{tab:ML_resultoverview}. Tables~\ref{tab:ML_result_kgramlowercase} and \ref{tab:ML_result_kgramoriginalcase} present the average accuracy and MAcc for each model across N-gram sizes ($N=2$ to $5$), reported separately for original-case and lowercase texts.
 
\begin{table} [tb] 
\centering
\small
\begin{tabularx}{\textwidth}{lCC}
\textbf{Model} & \textbf{Macro-accuracy (avg)} & \textbf{Accuracy (avg)} 
\\ \toprule 
SVM &	0.761 & 0.762 \\
LR &	0.701	& 0.725 \\
k-NN &	0.608	& 0.590 \\
DT & 0.632	& 0.631 \\
RF & 0.874 &	0.884 \\
ANN & 0.934	& 0.935 \\
\bottomrule
\end{tabularx}
\caption{\footnotesize
The average macro-accuracy and accuracy scores obtained for all six classification methods (SVM, LR, k-NN, DT, RF, ANN).}
\label{tab:ML_resultoverview}
\end{table}

\begin{table} [tb] 
\centering
\small 
\begin{tabularx}{\textwidth}{llCCCC}
\textbf{Model} &  & \textbf{2-gram} & \textbf{3-gram} & \textbf{4-gram} & \textbf{5-gram}
\\ \toprule 

SVM & MAcc avg&
0.328	& 0.869	& 0.934	& 0.899		
\\ & Acc avg& 0.309	& 0.874	& 0.933	& 0.911
\\ \midrule 
LR & MAcc avg&
0.502	&0.791	&0.795	&0.744		
\\ & Acc avg&  0.531	&0.807	&0.812	&0.770
\\ \midrule 
k-NN &  MAcc avg&
0.772	&0.603	&0.526	&0.559		
\\ & Acc avg&  0.765	&0.589	&0.502	&0.535
\\ \midrule 
DT &			 MAcc avg&					
0.635	&0.575	&0.666	&0.676		
\\ & Acc avg& 0.633	&0.581	&0.667	&0.675
\\ \midrule 
RF & MAcc avg&
0.919	&0.883	&0.857	&0.838		
\\ & Acc avg& 0.916	&0.893	&0.869	&0.857
\\ \midrule 
ANN & MAcc avg&
0.894	&0.945	&0.951	&0.954		
\\ & Acc avg&  0.899	&0.944	&0.951	&0.952
\\ 
\bottomrule
\end{tabularx}
\caption{\footnotesize
The average Macro-Accuracy (MAcc) and Accuracy (Acc) obtained for different N-gram sizes 
($N = 2$ to $5$), using fully lowercased text, for all six classification methods (SVM, LR, k-NN, DT, RF, ANN).
}
\label{tab:ML_result_kgramlowercase}
\end{table}

\begin{table} [tb] 
\centering
\small 
\begin{tabularx}{\textwidth}{llCCCC}
\textbf{Model} &  & \textbf{2-gram} & \textbf{3-gram} & \textbf{4-gram} & \textbf{5-gram}
\\ \toprule 
SVM	 & MAcc avg&				
            0.356	&0.872	&0.938	&0.894		
\\ & Acc avg&  0.348	&0.877	&0.938	&0.909
\\ \midrule 
LR	 & MAcc avg &
            0.510	&0.771	&0.777	&0.719		
\\ & Acc avg& 0.541	&0.793	&0.798	&0.748
\\ \midrule 
k-NN  & MAcc avg&
            0.755	&0.572	&0.511	&0.550		
\\ & Acc avg& 0.768	&0.550	&0.483	&0.526
\\ \midrule 
DT  & MAcc avg&
            0.611	& 0.577	&0.660	&0.657		
\\ & Acc avg& 0.603	&0.575	&0.662	&0.655
\\ \midrule 
RF  & MAcc avg&				
            0.914	&0.890	&0.848	&0.841		
\\ & Acc avg& 0.912	&0.898	&0.863	&0.863
\\ \midrule 
ANN  & MAcc avg&
            0.889	& 0.947	& 0.951	& 0.945	
\\ & Acc avg& 0.895	&0.943	&0.950	& 0.948
\\ 
\bottomrule

\end{tabularx}
\caption{ \footnotesize
The average Macro-Accuracy (MAcc) and Accuracy (Acc) obtained for different N-gram sizes 
($N = 2$ to $5$), using original case text, for all six classification methods (SVM, LR, k-NN, DT, RF, ANN).
}
\label{tab:ML_result_kgramoriginalcase}
\end{table}

\begin{figure} [tb] 
    \centering
    \subfloat[\centering]{\includegraphics[width=0.49\textwidth]
        {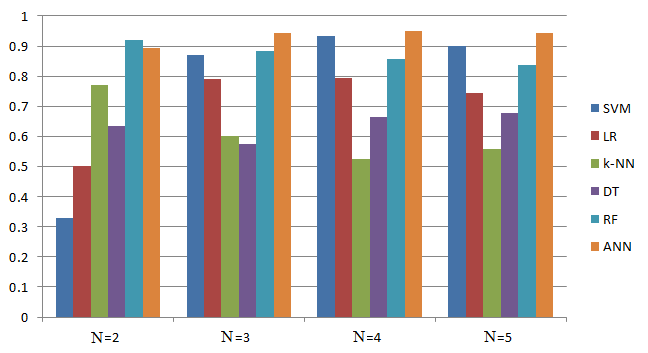}}
    \subfloat[\centering]{\includegraphics[width=0.49\textwidth]
        {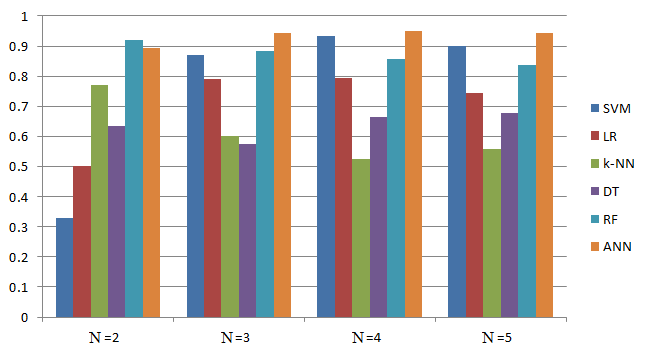}}
    \caption{\footnotesize
    Comparison of Macro-Accuracy (MAcc) scores for N-gram with $N = 2$ to $5$, for original case text (A), and fully lowercased text (B) and for all six classification methods (SVM, LR, k-NN, DT, RF, ANN). Values shown represent the average of the macro-accuracy scores computed over multiple experimental runs.}
     \label{fig:allKcases}
\end{figure}

ANN consistently outperformed other models, except at $k=2$, where RF 
achieved superior results. This trend is illustrated graphically in Figure~\ref{fig:allKcases}, which visualizes the MAcc values from the aforementioned tables.

For ANN, the highest performance was observed at $k=4$ using original-case letters, and at $k=5$ using lowercase letters—with lowercase slightly outperforming original case by 0.003 in MAcc. Conversely, RF achieved the best results at $k=2$, with average macro-accuracy values of 0.914 (original case) and 0.919 (lowercase).
Examining RF results further (Figures~\ref{fig:RFallKgr} )
we observe a slight decrease in average accuracy as $k$ increases, although MAcc values across different $k$ are not clearly separated.

\begin{figure}[tb] 
    \centering
        \centering
        \includegraphics[width=0.8\textwidth]{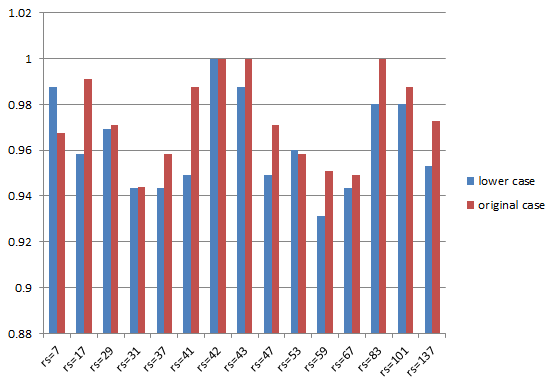} 
        \caption{\footnotesize The Macro-Accuracy (MAcc) scores corresponding to each of the 15 experimental runs of the Artificial Neural Network (ANN) and for the one split that produced the highest classification accuracy in the previous experiments. 
        \textbf{rs} denotes the $random\_state$ value used to initialize the ANN and shows the 15 distinct parameter settings.}
        \label{fig:ANN_all15}
\end{figure}

Notably, ANN achieved perfect classification (accuracy = 1.0) in one of the test runs.
This raises the question of whether such perfect accuracy reflects a consistent pattern for that particular train-test split, or if it is an isolated occurrence. To address this, we conducted an extended series of 15 experiments using different random seeds:

$ \{ 7,17,29,31,37,41,42,43,47,53, 59,67,83,101,137 \} $ 
\\for both original-case and lowercase texts, 
for that specific split
. As shown in Figure~\ref{fig:ANN_all15}, perfect accuracy was reached in four cases: twice for $random\_state=42$ (in both casing conditions) and two additional times for original-case texts.
The standard deviation of MAcc across these runs was approximately 0.02 (0.02039 for lowercase, 0.01938 for original case, and 0.02040 overall).
No clear pattern emerged relating accuracy variation to the random seed. However, original-case texts tended to perform slightly better on average, with a 0.01 higher MAcc (0.974 vs. 0.962 for lowercase), reversing the trend observed in Tables~\ref{tab:ML_result_kgramlowercase} and \ref{tab:ML_result_kgramoriginalcase}.
Given the small magnitude of these differences, and the reversed trend in the second set of experiments for ANN, we conclude that casing (original vs. lowercase) does not have a significant impact on classification performance.

\subsection{Comparison with Existing Results} 

In the literature, the best reported accuracies for authorship attribution often exceed 95\%. For example, Posadas et al. \cite{Posadas2017} combined word N-grams with the Doc2Vec method, achieving over 98\% accuracy on some test sets. Similarly, Zhang et al. \cite{zhang2015characterlevel} explored character-level convolutional neural networks (CNNs) and reported a minimal error rate of 1.31\% when using N-grams.

Reported accuracies vary widely depending on the dataset, feature selection, and classification methods, typically ranging from approximately 70\% upwards.

Authorship attribution for Romanian is still relatively underexplored, although recent interest has been increasing.
Notably, Nițu et al. \cite{nitu2024authorship} reported an F1-score of 0.87 on a 19-author dataset, improving to 0.95 on ROST, the same dataset used in our study. Also on ROST, Avram et al. \cite{avram2022comparison} achieved a lowest overall error rate of 20.40\% across several machine learning models, while Avram et al. \cite{avram2023bert} reported macro-accuracy up to 87\% using a Romanian pretrained BERT model.

Our approach, employing an Artificial Neural Network (ANN), achieved an initial average macro-accuracy of 0.935 across five data splits, with one split yielding perfect classification (macro-accuracy = 1.0). To evaluate the consistency of this result, we extended the analysis to 15 additional runs with randomized seeds (Section \ref{sec:bestresults}). For original-case texts, the expanded trials demonstrated a mean macro-accuracy of 0.974 (with a standard deviation $\sigma \approx 0.02$), aligning with the 0.979 baseline from initial testing on the same split.
Perfect classification occurred in four runs: once under lowercase transformation and three times with original casing. Notably, random seed 42 produced perfect classification under both text conditions.

These results demonstrate that our lightweight, character N-gram based approach with ANN matches or surpasses existing benchmarks for Romanian authorship attribution, offering a competitive alternative to more complex models such as BERT or transformer-based approaches.

\section{Conclusions and Future Work}
\label{sec:conclusions}
In this paper, we addressed the authorship attribution problem using a Romanian dataset previously employed in \cite{avram2023bert} and \cite{avram2022comparison}. To our knowledge, this is the first study to apply character N-gram based methods for Romanian authorship attribution.

We evaluated six machine learning techniques: Support Vector Machine (SVM), Logistic Regression (LR), k-Nearest Neighbor (k-NN), Decision Trees (DT), Random Forest (RF), and Artificial Neural Networks (ANN). Among these, the ANN model achieved the best performance, including perfect classification in four out of thirty runs for 5-gram features.

For future work, we plan to extend our investigation by incorporating additional feature types such as word N-grams, part-of-speech tags, and other linguistic markers. These enhancements aim to further improve attribution accuracy and deepen understanding of stylistic patterns in Romanian texts.


\begin{thebibliography}{1}

\bibitem{stamatatos2009survey}
Efstathios Stamatatos.
\newblock A survey of modern authorship attribution methods.
\newblock {\em Journal of the American Society for information Science and
  Technology}, 60(3):538--556, 2009.

\bibitem{potthast2011cross}
Martin Potthast, Alberto Barr{\'o}n-Cedeno, Benno Stein, and Paolo Rosso.
\newblock Cross-language plagiarism detection.
\newblock {\em Language Resources and Evaluation}, 45:45--62, 2011.

\bibitem{kestemont2014function}
Mike Kestemont.
\newblock Function words in authorship attribution. from black magic to theory?
\newblock In {\em Proceedings of the 3rd Workshop on Computational Linguistics
  for Literature (CLFL)}, pages 59--66, 2014.

\bibitem{neal2017surveying}
Tempestt Neal, Kalaivani Sundararajan, Aneez Fatima, Yiming Yan, Yingfei Xiang,
  and Damon Woodard.
\newblock Surveying stylometry techniques and applications.
\newblock {\em ACM Computing Surveys (CSuR)}, 50(6):1--36, 2017.

\bibitem{ngram2022}
Zheng Wanwan and Mingzhe Jin.
\newblock A review on authorship attribution in text mining.
\newblock {\em Wiley Interdisciplinary Reviews: Computational Statistics}, 15,
  04 2022.

\bibitem{koppel2009computational}
Moshe Koppel, Jonathan Schler, and Shlomo Argamon.
\newblock Computational methods in authorship attribution.
\newblock {\em Journal of the American Society for information Science and
  Technology}, 60(1):9--26, 2009.

\bibitem{dinu2008authorship}
Liviu~Petrisor Dinu, Marius Popescu, and Anca Dinu.
\newblock Authorship identification of romanian texts with controversial
  paternity.
\newblock In {\em LREC}, 2008.

\bibitem{avram2022comparison}
Sanda-Maria Avram and Mihai Oltean.
\newblock A comparison of several ai techniques for authorship attribution on
  romanian texts.
\newblock {\em Mathematics}, 10(23):4589, 2022.

\bibitem{avram2023bert}
Sanda-Maria Avram.
\newblock Bert-based authorship attribution on the romanian dataset called
  rost.
\newblock {\em arXiv preprint arXiv:2301.12500}, 2023.

\bibitem{nitu2024authorship}
Melania Nitu and Mihai Dascalu.
\newblock Authorship attribution in less-resourced languages: A hybrid
  transformer approach for romanian.
\newblock {\em Applied Sciences}, 14(7):2700, 2024.

\bibitem{kestemont2018overview}
Mike Kestemont, Michael Tschuggnall, Efstathios Stamatatos, Walter Daelemans,
  G{\"u}nther Specht, Benno Stein, and Martin Potthast.
\newblock Overview of the author identification task at pan-2018: cross-domain
  authorship attribution and style change detection.
\newblock In {\em Working Notes Papers of the CLEF 2018 Evaluation Labs.
  Avignon, France, September 10-14, 2018/Cappellato, Linda [edit.]; et al.},
  pages 1--25, 2018.

\bibitem{stamatatos2013robustness}
Efstathios Stamatatos.
\newblock On the robustness of authorship attribution based on character n-gram
  features.
\newblock {\em Journal of Law and Policy}, 21(2):421--439, 01 2013.

\bibitem{misini2024automatic}
Arta Misini, Ercan Canhasi, Arbana Kadriu, and Endrit Fetahi.
\newblock Automatic authorship attribution in albanian texts.
\newblock {\em Plos one}, 19(10):e0310057, 2024.

\bibitem{Howedi2014}
Fatma Howedi.
\newblock Text classification for authorship attribution using naive bayes
  classifier with limited training data.
\newblock 12 2014.

\bibitem{sapkota2015not}
Upendra Sapkota, Steven Bethard, Manuel Montes, and Thamar Solorio.
\newblock Not all character n-grams are created equal: A study in authorship
  attribution.
\newblock In {\em Proceedings of the 2015 conference of the North American
  chapter of the association for computational linguistics: Human language
  technologies}, pages 93--102, 2015.

\bibitem{boser1992training}
Bernhard~E Boser, Isabelle~M Guyon, and Vladimir~N Vapnik.
\newblock A training algorithm for optimal margin classifiers.
\newblock In {\em Proceedings of the fifth annual workshop on Computational
  learning theory}, pages 144--152, 1992.

\bibitem{fix1951discriminatory}
Evelyn Fix and Joseph~L. Hodges~Jr.
\newblock {Discriminatory analysis: Non-parametric discrimination: Consistency
  properties}.
\newblock Technical report, USAF School of Aviation Medicine, 1951.

\bibitem{fix1952discriminatory}
Evelyn Fix and Joseph~L. Hodges~Jr.
\newblock {Discriminatory analysis: Non-parametric discrimination: Small sample
  performance}.
\newblock Technical report, USAF School of Aviation Medicine, 1952.

\bibitem{altman1992introduction}
Naomi~S Altman.
\newblock {An introduction to kernel and nearest-neighbor nonparametric
  regression}.
\newblock {\em The American Statistician}, 46(3):175--185, 1992.

\bibitem{quinlan1986induction}
John~Ross Quinlan.
\newblock {Induction of decision trees}.
\newblock {\em Machine learning}, 1 (1986): 81-106.

\bibitem{zurada1992introduction}
Jacek~M Zurada.
\newblock Introduction to artificial neural systems, 1992.

\bibitem{Houvardas2006}
John Houvardas and Efstathios Stamatatos.
\newblock N-gram feature selection for authorship identification.
\newblock In {\em Artificial Intelligence: Methodology, Systems, Applications},
  2006.

\bibitem{2013Ramezani}
Reza Ramezani, Navid Sheydaei, and Mohsen Kahani.
\newblock Evaluating the effects of textual features on authorship attribution
  accuracy.
\newblock In {\em ICCKE 2013}, pages 108--113, 2013.

\bibitem{ngram-contin2006}
Yunita Sari, Andreas Vlachos, and Mark Stevenson.
\newblock Continuous n-gram representations for authorship attribution.
\newblock In Mirella Lapata, Phil Blunsom, and Alexander Koller, editors, {\em
  Proceedings of the 15th Conference of the {E}uropean Chapter of the
  Association for Computational Linguistics: Volume 2, Short Papers}, pages
  267--273, Valencia, Spain, April 2017. Association for Computational
  Linguistics.

\bibitem{Posadas2017}
Juan Posadas~Durán, Helena Gomez~Adorno, Grigori Sidorov, Ildar Batyrshin,
  David Pinto, and Liliana Chanona-Hernández.
\newblock Application of the distributed document representation in the
  authorship attribution task for small corpora.
\newblock {\em Soft Computing}, 21, 02 2017.

\bibitem{zhang2015characterlevel}
Xiang Zhang, Junbo Zhao, and Yann LeCun.
\newblock Character-level convolutional networks for text classification.
\newblock In C.~Cortes, N.~D. Lawrence, D.~D. Lee, M.~Sugiyama, and R.~Garnett,
  editors, {\em Advances in Neural Information Processing Systems 28}, pages
  649--657. Curran Associates, Inc., 2015.

\end{thebibliography}

\end{document}